# What Emotions Make One or Five Stars? Understanding Ratings of Online Product Reviews by Sentiment Analysis and XAI

Chaehan So[1]

[1]Information & Interaction Design,
Humanities, Arts & Social Sciences Division,
Yonsei University, 03722 Seoul, South Korea
`cso@yonsei.ac.kr`

**Abstract.** When people buy products online, they primarily base their decisions on the recommendations of others given in online reviews. The current work analyzed these online reviews by sentiment analysis and used the extracted sentiments as features to predict the product ratings by several machine learning algorithms. These predictions were disentangled by various methods of *explainable AI (XAI)* to understand whether the model showed any bias during prediction. Study 1 benchmarked these algorithms *(knn, support vector machines, random forests, gradient boosting machines, XGBoost)* and identified random forests and XGBoost as best algorithms for predicting the product ratings. In Study 2, the analysis of global feature importance identified the sentiment *joy* and the emotional valence *negative* as most predictive features. Two XAI visualization methods, *local feature attributions* and *partial dependency plots*, revealed several incorrect prediction mechanisms on the instance-level. Performing the benchmarking as classification, Study 3 identified a high *no-information rate* of 64.4% that indicated high *class imbalance* as underlying reason for the identified problems. In conclusion, good performance by machine learning algorithms must be taken with caution because the dataset, as encountered in this work, could be biased towards certain predictions. This work demonstrates how XAI methods reveal such prediction bias.

# 1    Introduction

Machine learning methods are known to be unrevealing of their learned knowledge due to their backbox nature. As they are frequently used to predict people's preferences to feed recommendation engines, it may be fruitful to scrutinize how they learned this prediction knowledge. To this aim, a new trend in artificial intelligence research caters to the emerging need for *interpretable or explainable AI (XAI)* [1, 2]. The present work applies such XAI methods to find out whether and how potential bias in models trained by machine learning algorithms can be detected and analyzed.

When people buy products online, they primarily base their decisions on the recommendations of others given in online reviews. To assess the usefulness of such user reviews, people rely on various factors such as the customer-rated helpfulness information [3]. To use the information contained in the review text for machine learning prediction, methods of *natural language processing (NLP)* have been applied. One popular NLP method, *sentiment analysis*, retrieves the emotional content in textual data. It has been applied to semantics like topics [4], aspects [5] or opinions [6], and domains including hotel reviews [7], movie reviews [8], or restaurant reviews [8].

The models of sentiment analysis differ in the sentiment categories. The most common category is *basic emotions*, a set of emotions from a theory by Ekman [9] believed to be universal in the sense of cross-cultural prevalence and inherited by all humans. The six basic emotions initially suggested by Ekman, namely happiness, surprise, fear, sadness, anger, disgust were validated by neuroimaging research [10, 11]. Another sentiment category is *emotional valence*, i.e. the binary categorization of emotion as positive or negative [12], also validated by neurocognitive research on event-related potentials (ERPs) [13] and fMRI [14].

Taken together, the preceding considerations lead to a research question that can be formulated as follows:

*How can XAI methods reveal potential bias in trained*
*machine learning models for the prediction of product ratings?*

To answer this research question, the current work analyzed Amazon online reviews by sentiment analysis and used the extracted sentiments as features to predict the product ratings by regression. These predictions, in turn, were disentangled by various XAI methods to understand whether the model showed any bias during prediction.

## 2 Method

The analysis was performed on a dataset on 28.332 consumer reviews on electronic products gathered by the data company Datafiniti on Amazon.com between February and April 2019. This dataset covered the complete range of electronic products branded by Amazon itself including the products Kindle, FireStick, and Echo.

### 2.1 Data PreProcessing

The data preprocessing encompassed the steps

a) removing technical information (id, dateAdded, dateUpdated, name, asins, imageURLs, keys, reviews.date, reviews.dateSeen, reviews.id, reviews.sourceURLs, reviews.username, sourceURLs)

b) removing data with nearly 100% missing values (e.g. reviews.didPurchase)

c) cleaning the levels of categorical variables from special characters (e.g. "&") and delimiting characters (e.g. ".", ","), and replacing spaces by underscores

These three steps resulted in a final sample size of n = 20238.

## 2.2 Descriptive Statistics

The variable *brand* showed 43.04% *AmazonBasics* and 56.96% *Amazon* occurrences.

In *primaryCategories*, 49.40% observations were in *Electronics*, whereas only 0.65% in *Electronics,Media* and 0.01% in *Electronics,Furniture* (the latter two categories were later merged with the former category), 42.60% in *Health & Beauty*, 5.92% in *Toys & Games,Electronics*, 1.36% in *Office Supplies,Electronics*, 0.03% in *Office Supplies*, 0.02% in *Animals & Pet Supplies*, and 0.01% in *Home & Garden*.

The variables for "recommend this product", *reviews.doRecommend*, number of helpful reviews, and *reviews.numHelpful* were present for only 12.200 of 28.332 observations in the dataset (56.8 %). This prevalence did not allow for the imputation of their missing values. The variable *reviews.doRecommend* was later removed after the prescreening as will be explained in the corresponding section. To allow the inclusion of the variable *reviews.numHelpful*, the missing values were set to 0.

After analyzing the content of all categorical variables, the following three had to be removed for subsequent analyses:

The variable *manufacturer* was removed because it contained essentially the same information as brand.

The variable *categories* was removed because it had very lengthy category levels that very difficult to grasp in meaning (e.g. the two most frequent category levels of 259 and 268 characters refereed to primary category *Health & Beauty*). Furthermore, the information was majorly overlapping with the variables *primaryCategories* and *asins*.

The variable *manufacturerNumber* had to be removed because it contained more than 53 class levels (65) which was incompatible with the machine learning libraries used in the benchmarking.

## 2.3 Design

The dataset was analyzed in three subsequent studies. Before these studies, a prescreening of the features was performed.

- Study 1 conducted model training as regression on the reviews-rating variable
- Study 2 applied methods of explainable AI (XAI) to analyze the best model on a global and local level
- Study 3 reframed the prediction as a classification task and converted the target variable into five class levels

## 2.4 Benchmarking Method

The present work compared the R implementations of the machine learning algorithms *knn* for k-nearest neighbors [15], *svmRadial* (method ksvm from library kernlab with radial kernel) [16] for support vector machines, *rpart* for CART decision trees [17], *rf* (library randomForest) [18] for Random Forests, *gbm* [19] for Gradient Boosting Machines and *xgbTree* [20] for XGBoost on decision trees.

## 2.5 Explainable AI (XAI)

Interpretability, in the machine learning context, is defined as the "ability to explain or to present in understandable terms to a human" [21].

The emerging need for interpretability of machine learning algorithms has led to the new branches of *interpretable AI* which has recently been referred to as *explainable AI (XAI)*. This trend can be witnessed at the most esteemed machine learning conferences in recent years. In 2017, Been Kim and Finale Doshi-VelezI published a paper [21] and offered a tutorial on *interpretable machine learning* at ICML, while Neurips had a same-named tutorial in the same year. In 2018, both conferences offered workshops that contained the term *explainability* in the titles.

Along with the explosive growth of explainable AI research, a multitude of programming libraries has become available. For Python, the popular python distribution platform *pip* has offered the libraries *xai. yellowbrick, ELI5, lime, MLextend*, and *SHAP*. For the R programming language, R package distribution platform *CRAN* has offered the libraries *breakdown, DALEX, lime, modelDown, pdp*, and *shapper*, among others. Although this list is not extensive, it shows already the interest of the machine learning community to develop programming tools to support explainable AI analysis.

The XAI methods applied in the present work shall be defined in the following.

**Feature importance**

The variable or feature importance is a numeric value of a feature' prediction impact, i.e. its global relevance for generating the prediction in the trained model. This relevance is model-dependent, i.e. for tree-based models like gradient boosting machines or random forests, it corresponds to its role in splitting the trees, whereas for linear models, it corresponds to the normalized regression coefficient.

**Local attributions**

The prediction of a trained model can be visualized to show each feature's local attributions to the averaged prediction of a few unseen observations. Such visualization can serve to uncover the local, i.e. instance-level, role of a feature that typically differs a lot between observations. In other words, a feature's variance, which is hidden by the analysis of global feature importance, can be understood by this feature's local attributions' visualization.

**Partial dependency plot**

The partial dependency plot (pdp) displays the marginal effect of several features on the target variable of a trained machine learning model [22]. The partial dependency is the degree to which the target variable partially depends on each feature individually, i.e. partially, and over its whole range.

All of the above three XAI methods may serve not only to understand the functioning of a trained model, but importantly, may also serve to make important decisions for model improvement. Features may e.g. be discarded due to low feature importance, overly high variance in local attributions, or wrong or weak relationships uncovered in partial dependency plots.

## 3 Results

### 3.1 Prescreening

Before the main studies, the current work performed a pre-screening of the preprocessed dataset's features on approximately half the dataset.

The prescreening encompassed the analysis of feature importance that showed by far the highest importance for the variable *reviews.doRecommend*. It is created when users (who have written their online review on Amazon.com) answer *Yes* or *No* to the statement *I recommend this product*. This variable had to be removed for two reasons:

First, it was only present in 56.8% of the dataset. Therefore, on the one hand, its missing values could not be imputed, and on the other hand, its inclusion as a feature would have dramatically reduced the training set by almost half its size.

Second, this variable is too obviously related with the target variable, the product rating. Answering *Yes* to a product recommendation is semantically a strong endorsement equivalent with a product rating of five stars. This strong relationship was further evidenced by transforming it into a numeric variable which revealed a very high correlation (r = .648) with the target variable. Since the focus of the current work was on the predictive power of NLP, the variable *reviews.doRecommend* had to be removed to enable the comparison of feature importance between sentiments.

### 3.2 Study 1

The benchmarking results for the regression task are displayed in Figure 1. Random forests yield the best prediction (lowest RMSE), whereas support vector machines with linear kernel yield the worst prediction.

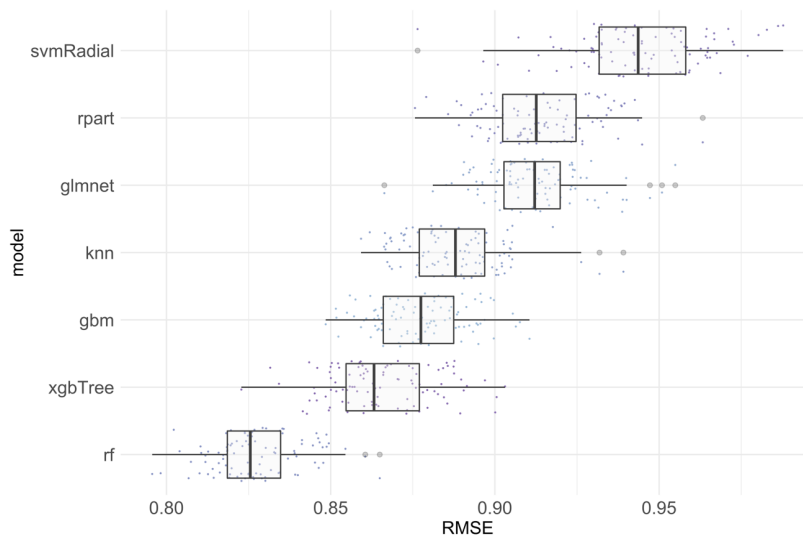

**Fig. 1.** Benchmarking Results Regression, n = 20238

The benchmarking results do not yield any explanatory value other than the algorithms' ranking. This black box characteristic was an expected result, and demonstrates the need for new methods to explain how the predictions were generated by the trained model. Such methods are provided by XAI methodology and shall be applied in the subsequent studies.

### 3.3 Study 2

The goal of Study 2 was to explain by XAI methods how the trained model of the best algorithm from Study 1's benchmarking, random forests, calculates its prediction. The explanation is performed both on the global and on the local level.

On the global level, the variable or *feature importance* can be retrieved from the trained model. As can be seen in Figure 2, the visualization of feature importance reveals that the basic emotions *joy, trust, fear* and *anticipation,* as well as *negative* emotional valence yield the highest predictive value for random forests. In contrast, the *positive* emotional valence scores among the lowest, and the categorial variable *primaryCategories* shows the lowest feature importance.

Even though these feature importance scores reflect the overall importance of each feature, it should not disguise the fact that their prediction influence varies across all observations. This variance also means that an individual feature's impact on the target variable may reflect very differently in one observation from another.

Nevertheless, in the presence of many features, the feature importance scores can be utilized to shrink the feature set reasonably. In most cases, it would be pragmatic to discard features with feature importance scores lower than 10. This would concern features that have lower prediction impact than 10% of the most relevant feature.

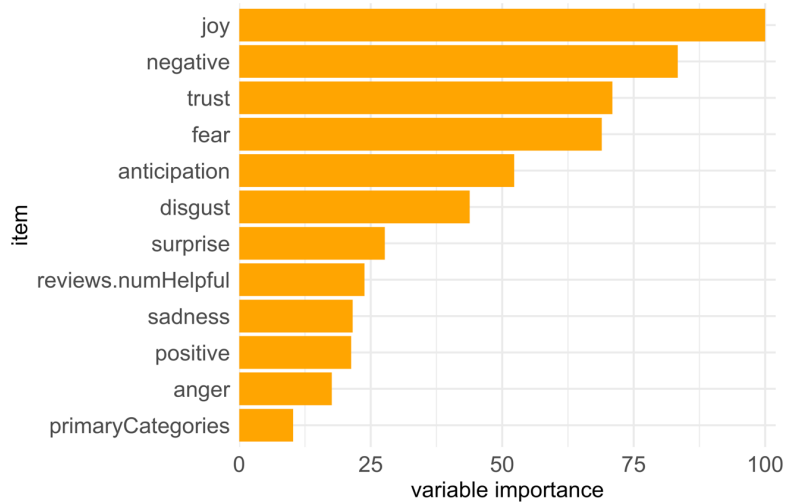

**Fig. 2.** Global Feature Importance, random forests, n = 20238

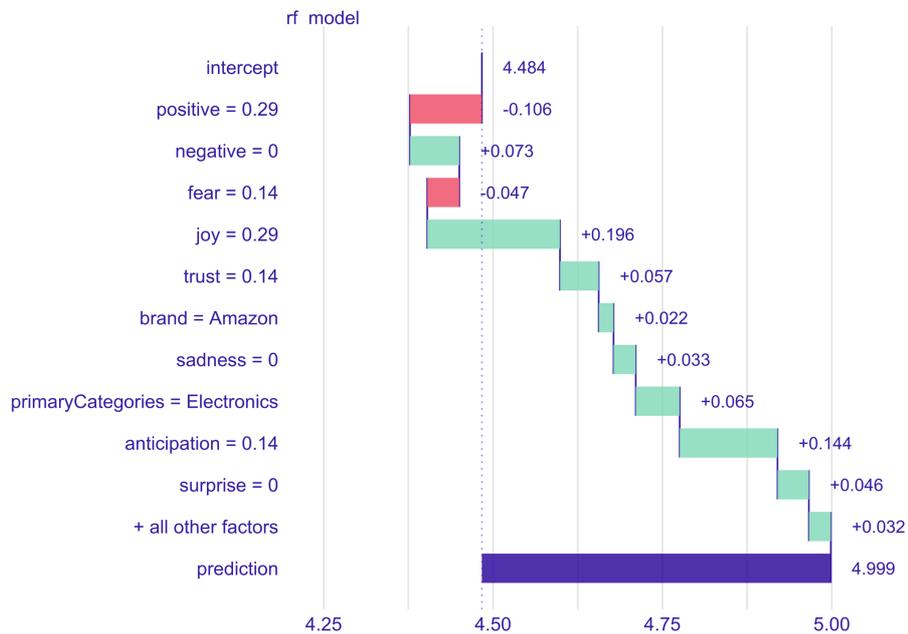

**Fig. 3.** Local Attributions of Features, random forests, n = 6

On the local level, the prediction of the trained random forests model was computed over six unseen observations.

The local attributions diagram in Figure 3 shows that the prediction of 4.999 for the reviews-rating was mainly negatively influenced by low values of *positive* and *fear*. On the other side, low levels of all the remaining variables contributed positively to the prediction. Apart from that, although *trust* and *anticipation* had the same score from sentiment analysis (both 0.14), their local attribution was inverse to their global feature importance – *trust* with higher feature importance had a lower local attribution (+0.057) than *anticipation* (+0.144).

Generally speaking, it was not evident why the overall low emotional scores should lead to a prediction (4.999) near the maximum value (5.0). Furthermore, it was not plausible why a positive level of *positive* valence (0.29) should have a negative impact. Other mentionable aspects were the extremely high intercept (4.484) as well as the relative irrelevance of all other factors (with a contribution of +0.032 to the prediction of 4.999).

Taken together, it was not plausible why some features that should have a positive contribution had a negative one (positive valence), and features that had stronger feature importance had a lower local attribution. This puzzle required a more detailed analysis on the instance level.

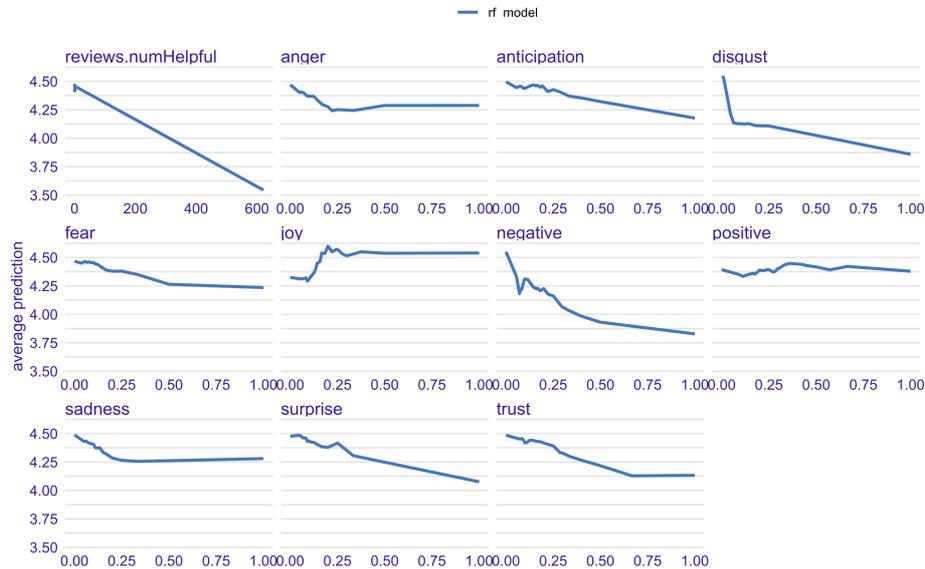

**Fig. 4.** Partial Dependency Plot on Local Observations – Random Forests

Apart from the local attribution analysis, another useful instance-level method is the analysis of *partial dependency*. This method can reveal the contribution to the prediction across the range of each feature.

The partial dependency plots (Figure 4) detected systematic weaknesses, i.e. the algorithm did not correctly identify the relationships between several variables – e.g. *anticipation* and *trust* were incorrectly identified to have a negative relationship, Furthermore, although *anger, fear*, and *sadness* were correctly identified as negatively related to the target variable, the relationship was very weak and should be stronger. Besides, the variable *positive* (emotional valence) should have a strong relationship with the target but was identified to be not related at all (zero slope).

### 3.4 Study 3

To further understand the algorithm's functioning, the prediction task was reformulated as a classification task. This reformulation was permissible because the target variable, the reviews-rating, contained only integer values that were directly convertible into five classes with each of the five stars' rating as class level.

The benchmarking, visualized in Figure 5, shows a similar result as Study 1. Again, random forests yield the best prediction, indicated by the highest accuracy on the hold-out set of 72.9%. Nevertheless, the *no-information rate* showed a high value of 64.4%. This reveals a severe class imbalance in the dataset that explains the partially wrong relationships identified in the previous studies.

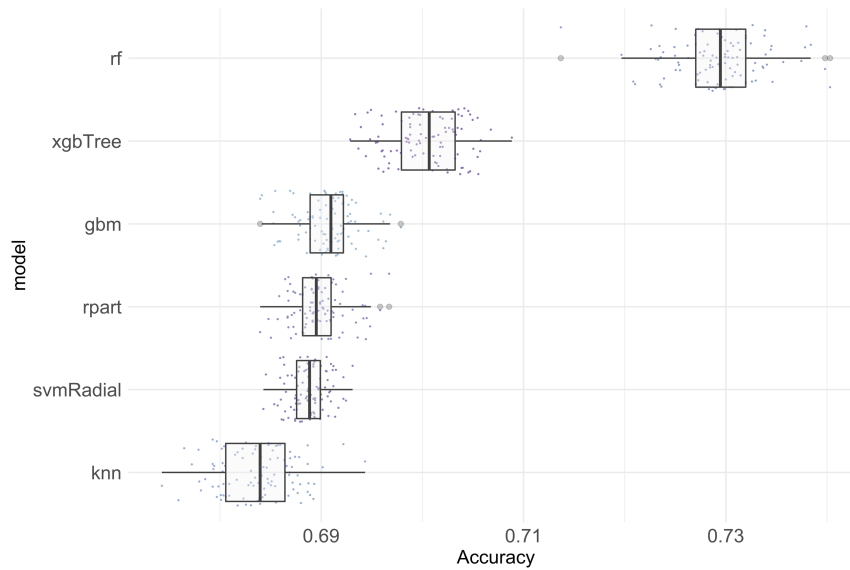

**Fig. 5.** Benchmarking Results Classification, n = 20238

## 4    Discussion

Online companies often use machine learning to make predictions on customer behavior and provide recommendations to users. The present work showed that such predictions can be severely biased, and that a global analysis of feature importance is insufficient to detect such bias. Only the analysis of local feature attributions and partial dependency plots could reveal that the predictions were severely biased towards positive ratings.

### Acknowledgment

This research was supported by the Yonsei University Faculty Research Fund of 2019-22-0199.